\pdfoutput=1
\documentclass[11pt]{article}

\usepackage[preprint]{acl}
\usepackage{pgf}
\usepackage{times}
\usepackage{latexsym}

\usepackage[T1]{fontenc}
\usepackage[utf8]{inputenc}

\usepackage{microtype}

\usepackage{inconsolata}

\usepackage{graphicx}

\title{Bonafide at LegalLens 2024 Shared Task: Using Lightweight DeBERTa Based Encoder For Legal Violation Detection and Resolution}

\author{Shikha Bordia \\
  \texttt{bordiashikha06@gmail.com} \\}

\begin{document}
\maketitle
\begin{abstract}

In this work, we present two systems---Named Entity Resolution (NER) and Natural Language Inference (NLI)---for detecting legal violations within unstructured textual data and  for associating these violations with potentially affected individuals, respectively. Both these systems are lightweight DeBERTa based encoders that outperform the LLM baselines. The proposed NER system achieved an F1 score of 60.01\% on Subtask A of the LegalLens challenge, which focuses on identifying violations. The proposed NLI system achieved an F1 score of 84.73\% on Subtask B of the LegalLens challenge, which focuses on resolving these violations by matching them with pre-existing legal complaints of class action cases. Our NER system ranked sixth and NLI system ranked fifth on the LegalLens leaderboard. We release the trained models and inference scripts\footnote{\url{https://github.com/BordiaS/LegalLens_inference}}. 

\end{abstract}

\section{Introduction}

%The widespread use of the internet has led to a vast amount of unstructured textual data, making it challenging to detect legal violations that may be concealed within this information. 

Social networks and other online platforms are increasingly becoming effective tools to address consumer complaints; however, the vast amount of unstructured textual data makes it challenging to identify valid complaints and if they are associated with any legal violations. There is a pressing need to develop sophisticated methods to identify these hidden breaches, as they have significant implications for individual rights and legal obligations, if any. 

In this regard, \citet{bernsohn2024legallens}  propose two subtasks
%that can serve as components to identify legal violations and associate them with resolved class action cases
---Subtask A, Legallens NER (Named Entity Recognition), to detect legal violations mentioned in the text and Subtask B, Legallens NLI (Natural Language Inference), to match the detected violations with resolved class action cases.
% abiding by the court rulings while maintaining a fair unbiased claims handling framework is now much more crucial. Taking a proactive approach to handling these challenges benefits companies in building a positive image and also protects them from reputational and litigation risks. 
%The proposed tasks in the 2024 
%\citet{bernsohn2024legallens} propose a set up of two tasks that can serve as components to identify these violations using and associate these violation by matching them with resolved class action cases. They propose Subtask A: LegallensNER to detect and Subtask B: LegallensNLI to resolve violations.
To address these subtasks in this paper, we propose NER and NLI models based on training DeBERTaV3 encoders. We finetune task-specific encoders on our synthetically augmented dataset. In summary, we list our findings here:

\begin{enumerate}
    \item Continuing to pretrain an already powerful general domain task-specific model on our subtask can boost the  performance of our system.

    \item While synthetic data can significantly boost the capabilities of models, it's crucial to recognize that surpassing specific thresholds of training data volumes may not necessarily result in proportional enhancements in performance.

     \item Scaling laws suggest that Large Language Models (LLMs) show predictable  performance improvements. However, smaller models can either match or perform better using appropriate training objectives and data, specifically for classification tasks.
    
    %Synthetic data can substantially enhance the performance of the models %due to limited size of annotated data.
    %However, it is important to note that exceeding certain threshold or training data size may not lead to proportional improvements in performance
\end{enumerate}
%demonstrates that the forementioned task is achievable with the latest research advancements in Legal NLP as well as social media NLP?? (name a few). Employing such models and techniques is also more viable now with reducing costs and improving accuracy. 

%1. Present your research topic
%Introduce the broad subject area and provide relevant background information to contextualize your study.
%Use a hook to capture the reader's interest in the first sentence.
%2. Summarize existing research
%Briefly review the most relevant published literature on your topic.
%Identify gaps, limitations or areas that need further investigation.
%3. State your research problem and objectives
%Clearly define the specific research problem, question or hypothesis you aim to address.
%Explain the significance and potential outcomes of your study.
%4. Describe your approach
%Provide a brief overview of your methodology.
%Explain how your approach differs from or builds upon previous research.
%5. Give an overview of the paper's structure
%Outline the key sections and content covered in th

In Section \ref{related_work}, we examine the related works on NER and NLI tasks. Section \ref{methods} provides an overview of the methodologies employed for the tasks.  In Section \ref{experiments}, we describe the experimental setup. Section \ref{results} discusses the results and findings; Section \ref{conclusion} discusses the conclusions. %and future research lines.

\section{Related Works}
\label{related_work}
\paragraph{NER}

Research in NER has evolved from statistical models such as Maximum Entropy \citep{borthwick1998nyu}, Hidden Markov Models \citep{bikel1999algorithm}, and Conditional Random Fields (CRF)\citep{mccallum-li-2003-early}, using bidirectional RNNs, often combined with CRF layers \citep{huang2015bidirectional, ma-hovy-2016-end, lample-etal-2016-neural} to using transformer-based models \citep{vaswani2017attention}. This transition has enabled accurate and robust entity recognition across various domains and languages. In legal domain, variations of BERT-based transformers \citep{devlin2018bert} like RoBERTa \citep{liu2019roberta}, DeBERTaV3 \citep{he2021debertav3}, LegalBERT \citep{chalkidis2020legal}, %and ALBERT, and implemention of XLNet, and Legal-LUKE 
with BiLSTM and CRF layers on the top \citep{huo-etal-2023-antcontenttech,ningthoujam2023researchteam_hcn} have given state-of-the-art performance on legal NER tasks \citep{kalamkar2022named, modi2023semeval}. Legallens NER task has four sets of entity types that have not been previously explored in legal NER research. %as described in Section \ref{ner_task}. 
In this work, we use the recently proposed DeBERTaV3 based GLiNER \citep{zaratiana2023gliner} architecture that  outperforms both ChatGPT \citep{NEURIPS2020_1457c0d6} and fine-tuned LLMs in zero-shot evaluations on various NER benchmarks. %The architecture combines global linearization and embedding representations.

\begin{table}[t]
\scriptsize
\centering
\caption{The distribution of number of words by entity type in the LegalLens NER training dataset}

\begin{tabular}{lcccc} 
%\toprule
   \hline
\textbf{LAW} & \textbf{VIOLATED BY} & \textbf{VIOLATED ON}& \textbf{VIOLATION} \\
   \hline
%\midrule
 4.14    &  2.19       &    3.24       & 12.39  \\
%$77.57_{\pm 1.35}$ & $59.06_{\pm 0.55}$ & $76.88_{\pm 2.06}$ & $62.83_{\pm 2.57}$ \\
%\bottomrule
\end{tabular}
\label{fig:number_of_words}
\end{table}

% \begin{table}[ht]
% \scriptsize
%   \centering\textbf{}
%   \toprule
%   \begin{tabular}{lcccc}
%     \hline
%     % \textbf{LAW} & \textbf{VIOLATED BY} & \textbf{VIOLATED ON} & \textbf{VIOLATION} \\
%     \texttt{LAW} & \texttt{VIOLATED BY} & \texttt{VIOLATED ON} & \texttt{VIOLATION} \\
%         \midrule
    
%         \hline
    
%     4.14    &  2.19       &    3.24       & 12.39  \\
%     \bottomrule
%   \end{tabular}
% \caption{Example commands for accented characters, to be used in, \emph{e.g.}, Bib\TeX{} entries.}
%   \label{tab:NER_by_ents}
% \end{table}

% \begin{figure}
%     \includegraphics[width=\columnwidth]{no_of_words.png}
%     \caption{The distribution of number of words by entity type in the LegalLens NER training dataset }
%     \label{fig:number_of_words}

% \end{figure}

\paragraph{NLI}

NLI classifies the logical relationship between a premise (a given statement) and a hypothesis (a proposed conclusion) as entailed, contradictory, or neutral. Early work on NLI focused on rule-based systems and logical inference \citep{giampiccolo2007third}.  The advent of large-scale datasets, such as the SNLI \citep{bowman2015large}, MultiNLI corpus \citep{williams2017broad}, XNLI \citep{ conneau2018xnli} enabled the development of sophisticated models. %Neural network architectures showed significant success in NLI tasks. \citep{parikh2016decomposable,chen2016enhanced} %introduced a decomposable attention model, while Chen et al. (2017) proposed the Enhanced Sequential Inference Model (ESIM), both achieving state-of-the-art results at their time.
Transformer-based models such as RoBERTa \citep{liu2019roberta} and XLNet\citep{yang2019xlnet} have pushed the limits of NLI performance by giving human-like scores.

%Recent research has focused on improving the robustness and generalization of NLI models. McCoy et al. (2019) highlighted the tendency of models to rely on superficial cues rather than true understanding. This has led to efforts in creating more challenging datasets and developing models that can better capture the nuances of language and reasoning.

%Cross-lingual NLI has also gained attention, with datasets like XNLI (Conneau et al., 2018) facilitating research on multilingual inference. This area explores how NLI models can generalize across languages and transfer knowledge between them.
NLI is a critical task in NLP that serves as a benchmark for natural language understanding. Although significant progress has been made, challenges remain in developing systems that can perform robust and generalizable inference across diverse domains and languages. %It is advisable to use domain specific finetuned smaller models than larger general purpose models. 
In this work, we use Tasksource's NLI model and finetune it on the LegalLens NLI dataset.  Tasksource is a framework that harmonizes data sets for multitask learning and evaluation in NLP by providing a collection of pre-processing methods \citep{sileo-2024-tasksource}.

%LegalLens NLI task facilitates the correlation of multiple unstructured text associated with the same violation, thereby enabling the matching of extracted violations identified by the NER task with pre-existing legal complaints of class action cases

%NLI facilitates the correlation of multiple unstructured text associated with the same violation, thereby enabling the matching of extracted violations identified by the NER task with pre-existing legal complaints of class action cases

\paragraph{Data Augmentation}

The advent of LLMs has introduced a novel approach to data augmentation in machine learning tasks \citep{he2021generate, gan2019improving, hosseini2024synthetic}. Leveraging the capabilities of these models, we employ two distinct strategies to enhance our datasets. For the NER task, we utilize few-shot learning techniques to expand the existing dataset. This method allows us to generate additional, contextually relevant examples based on a small number of initial samples. Concurrently, for the NLI dataset, we implement a paraphrasing approach. This technique involves reformulating the sentences---premise and hypothesis---while preserving their semantic content, thereby increasing the diversity and robustness of our training data. This approach also preserves the original label distribution. We use Mixtral 8x7B model \citep{jiang2024mixtral}, a state-of-the-art LLM, to augment both the datasets. The specific prompts used for these augmentation tasks are detailed in the Appendix  \ref{appendix} for both the subtasks, ensuring transparency and reproducibility of our methods.

%LLMs have opened the possibility of generating synthetic data to assist machine learning tasks
%\citep{he2021generate,  gan2019improving, hosseini2024synthetic}.
%\citep{he2021generate}, \citep{gan2019improving} \citep{hosseini2024synthetic}.
%We use few-shot learning to augment the NER dataset and paraphrasing to enhance the NLI dataset using Mixtral 8x7B \citep{jiang2024mixtral}. Prompts provided in the Appendix \ref{appendix}.
%LLMs excel at paraphrasing  \citep{yadav2024pag}. We use 
\begin{table*}[t]
  \centering
  \begin{tabular}{lccc}
    \hline
    \textbf{Model} & \textbf{Precision} & \textbf{Recall}  & \textbf{F1}  \\
    \hline
    \verb|gliner_small-v2.1|   &  70.26 & 45.83   & 55.47           \\
    \verb|gliner_base|    &  71.30  & 47.02  &   56.67      \\
    \verb|gliner_small|  & 72.32  & 45.71   & 56.02           \\
      \verb|gliner-bi-base-v1.0|    & \textbf{83.30} & 46.31   & \textbf{59.53}  \\
   \verb|gliner-bi-small-v1.0|    & 74.00 & 48.39   & 58.52 \\
    \verb|gliner-poly-small-v1.0|    & 71.04 & \textbf{49.64}   & 58.44 \\
  \end{tabular}
\caption{
Comparison of different GLiNER architectures on LegalLens NER development dataset. The table showcases the models and their respective performance }%\texttt{gliner\_small-v2.1}}
  \label{tab:NER_by_model}
\end{table*}

\section{Methodology}
\label{methods}
In this section, we introduce our approach for each of the subtasks.

\subsection{Subtask A: LegalLens NER }
\label{ner_task}
\paragraph{
Problem Statement
}
The NER task aims to detect legal violations in social media posts and online reviews. %The dataset is categorized by Cause of Action that refers to a set of legal reasons that justify the right to seek legal remedy. There is no overlap between the training and test sets provided to prevent any data leakage while training the model. 
The training and development datasets consist of 710 and 617 data points. We specifically identify the following entities:  \texttt{LAW} (law or regulation breached), \texttt{VIOLATION} (content describing the violation), \texttt{VIOLATED BY}(entity committing the
violation) and \texttt{VIOLATED ON} (victim or affected party).The average number of words range between 2.19 and 4.14 for \texttt{LAW}, \texttt{VIOLATED BY} and \texttt{VIOLATED ON} while the average number of words for \texttt{VIOLATION} is 12.39  as shown in  Table \ref{fig:number_of_words}. 

%\paragraph{Data Exploration}

\paragraph{Contribution}
Our main contributions are as follows:
\begin{enumerate}
\item[\textbullet]We finetune a lightweight bidirectional transformer encoder GLiNER proposed by \citet{zaratiana2023gliner}, that uses DeBERTaV3  \citep{he2021debertav3} as backbone. It is trained on Pile-NER dataset \citep{zhou2023universalner}.

\item[\textbullet]We experiment with %comprehensive analysis on the performance of 
the architectures---single, bi-encoder and polyencoder---proposed by \citet{zaratiana2023gliner}
\end{enumerate}

 All the pre-trained checkpoints of these
models are taken from the Hugging Face hub repository.

\subsection{Subtask B: LegalLens NLI }

\paragraph{Problem Statement} %The objective of the NLI task is to match resolved class action cases with the violations identified by the NER model. The premise consists of the summary of news articles from a legal news websites. The hypothesis consist of synthetically generated social media posts that mimic common life situations of potential legal violations. The data set consists of 312 data points in four legal domains : Consumer Protection, Privacy, TCPA and Wage. The models are trained on three domains and tested on the fourth one to avoid any potential risk of data leakage. 

The NLI task aims to link resolved class action cases with violations detected by the NER model. The premise comprises summaries of legal news articles, while the hypothesis consists of synthetically generated social media posts that mimic potential legal violations. The dataset includes 312 data points across four legal domains: Consumer Protection, Privacy, TCPA, and Wage. 
\paragraph{
Contribution
}
\begin{enumerate}
\item[\textbullet]We finetune a multitask DeBERTaV3 based encoder, Tasksource \citep{sileo-2024-tasksource},  that casts all the classification tasks as natural language inference and trains the model on 600+ English tasks simultaneously to achieve state-of-the-art performance at its size.

\item[\textbullet] 
We propose synthetic data generation to enhance the performance of the model. We employ Mixtral 8x7B by \citet{jiang2024mixtral} to generate paraphrases for each premise-hypothesis pair. The class labels (Entailed, Contradict, and Neutral) remain unchanged. This approach doubles the size of the training data  while preserving the original class distribution.

\item[\textbullet] 
Augmenting the NLI dataset boosted the final F1 score metric by a significant margin of 7.65\%. 
\end{enumerate}

%\paragraph{Prompt} %We use Mixtral 8x7B by  to generate paraphrases of each pair of premise and hypothesis. The class label (Entailed, Contradict and Neutral) is kept as is. This enriches the training data two folds while keeping the class weights as original. 

\section{Experimental Settings}
\label{experiments}

%\subsection{NER}
\paragraph{NER}
 We finetune the GLiNER models on the LegalLens NER dataset using a dropout rate of 0.5 and a batch size of 8. We employ AdamW optimizer with a base learning rate of 1e-5 for pre-trained layers (the transformer backbone, DeBERTaV3) and 5e-5 for non-pre-trained layers (FFN layers, span representation).  The model is trained to a maximum of 10 epochs, starting with a 10\% warm-up phase, followed by a decay phase using a linear scheduler. We save the best checkpoint and, subsequently, reduce the learning rate to 5e-6, and train this checkpoint until convergence. To address class imbalance, we use focal loss, instead of cross-entropy loss, with alpha 0.75 and gamma 2. 
 
We experiment with three different architectures proposed by \citet{zaratiana2023gliner} and Knowledgator 
Engineering\footnote{\href{https://blog.knowledgator.com/meet-the-new-zero-shot-ner-architecture-30ffc2cb1ee0}{Knowledgator Blog link}}---original GLiNER, the bi-encoder and the poly-encoder  as shown in Table \ref{tab:NER_by_model}. 
During inference, we utilize a model threshold of 0.8 to compute performance metrics. Additionally, we implement a rule to eliminate false positive entities. In the event that multiple entities of the same type are extracted, we discard the entity with the lowest confidence score and retain the one with the highest score. This approach resulted in an improvement in the F1 score by 0.5\%, reaching 60.01\%. 
%\paragraph{Inference}

% \begin{figure}
%     \includegraphics[width=\columnwidth]{cm.png}
%     \caption{Final NLI Confusion Matrix on the test dataset }
%     \label{fig:cm}

% \end{figure}

%\subsection{NLI}

\begin{table}
  \center\textbf{}
  \begin{tabular}{lccc}
    \hline
    \textbf{Entity Type} & \textbf{Precision} & \textbf{Recall}  & \textbf{F1}  \\
    \hline
    \texttt{LAW}  &  73.40 & 92.00   & 81.66           \\
    \texttt{VIOLATED BY}    & 88.16  & 89.33   &   88.74      \\
    \texttt{VIOLATED ON}     & 71.43  & 73.33   & 72.37           \\
  \texttt{VIOLATION}     & 68.17 & 39.29   & 49.85  \\
   \hline
   \texttt{micro avg}     & 71.93 & 51.49   & \textbf{60.01}  \\
 % \verb|macro avg|     & 75.29 & 73.49   & \textbf{73.15} \\   
  \end{tabular}
\caption{Entity level metrics of the best performing model \textbf{gliner-bi-base-v1.0} integrated with predefined rules}
  \label{tab:NER_by_ents}
\end{table}

\begin{table*}[ht]
  \centering
  \begin{tabular}{lcccccr}
    \hline
    \textbf{Model} & \textbf{Consumer Protection} & \textbf{Privacy}  & \textbf{TCPA} & \textbf{Wage} &  \textbf{Macro F1}\\
    \hline
  
    \verb|tasksource| (orignal) &85.48\    &  76.07  &  62.16  & 81.56 & 76.31   \\ %77.08
    %\verb|tasksource| (org) &85.48\    &  76.07  &  62.16  & 84.61 & 77.08   \\ %77.08
    %\verb|Data Augmentation + Finetuning|     & 86.65 & 74.95  & 79.82   & 100      \\
      \verb|tasksource| (augmented)  & \textbf{88.71} & \textbf{85.88}  & \textbf{79.72}   & \textbf{84.61}    & \textbf{84.73} \\  %84.73

  \end{tabular}

  \caption{
  Comparison of Tasksource's model performance on LegalLens NLI's dev dataset. %The first row shows the performance on the original dataset.
  The second row shows the improved performance using the augmented dataset  over the original dataset as the training data by 7.65\%}
  \label{tab:NLI_table}
\end{table*}
\paragraph{NLI}
We train four models and test them on each legal domain. Each of these four models is trained on three domains at once and tested on the fourth to prevent data leakage as described by \citet{bernsohn2024legallens}. For each domain, we finetune %\texttt{sileod/deberta-v3-small-tasksource-nli}  and \texttt{sileod/deberta-v3-base-tasksource-nli} 
Tasksource's NLI model using a learning rate of 2e-5, a sequence
length of 256, and a batch size of 8 for a maximum of 7 epochs using a cosine scheduler. We save the best checkpoint and
reduce the learning rate to 2e-6, and further train it until convergence. 
As shown in Table \ref{tab:NLI_table}, the synthetically augmented dataset boosted the performance of the models on the development dataset by 7.65\%.  
%\paragraph{Inference}

% \begin{table}
%   \center\textbf{}
%   \begin{tabular}{lccc}
%     \hline
%     \textbf{Model} & \textbf{First-class} & \textbf{Second-class A}  & \textbf{Second-class B}  \\
%     \hline
%     \texttt{baseline }  &  1 & 21   & 18           \\
%     \texttt{tasksource}    & 2  & 15  &   30      \\

%  % \verb|macro avg|     & 75.29 & 73.49   & \textbf{73.15} \\   
%   \end{tabular}
% \caption{Entity level metrics of the best performing model \textbf{gliner-bi-base-v1.0} integrated with predefined rules}
%   \label{tab:NER_by_ents}
% \end{table}

\section{Results and Discussions}
\label{results}
\paragraph{NER}

The original GLiNER architecture employs bi-directional encoder.   The entity labels, separated by [SEP] token, and  the input sequence are concatenated and then passed through the encoder model. The bi-encoder architecture decouples the entity labels and input sequence. The poly-encoder uses fuses the entity label and input sequence encoder representations together to capture the interactions between them. The bi-encoder model, \texttt{gliner-bi-base-v1.0}, has best performance with an F1 score of 59.53\% and the highest precision of  83.30\%. The polyencoder model, \texttt{gliner-poly-small-v1.0}, gave the highest recall of 49.64\% as shown in Table \ref{tab:NER_by_model}.

%we observe that the bi-encoder model outperforms both the original GLiNER and the poly-encoder model. 

Our experiments reveal that shuffling entity order and randomly dropping entities did not affect the metrics. After identifying the best model, we trained it on a synthetic dataset generated using few-shot learning. %Prompts and examples are provided in Appendix \ref{appendix}. 
However, this approach did not yield any improvement in results. We then applied rule-based entity filtering, which improved the development dataset results by 0.5\%, increasing the final F1 score from 59.53\% to 60.01\%. The system ranked sixth on the leaderboad with an F1 score of 33.00\% on the test dataset \citep{hagag2024legallenssharedtask2024}.

Table \ref{fig:number_of_words} illustrates the distribution of word count by entity type. The \texttt{VIOLATION} entity type averages 12.39 words, compared to a maximum of 4.14 for the other three types, increasing the complexity of the task. The model performs better on shorter entities, as shown in Table \ref{tab:NER_by_ents}. Previous research has shown that NER models struggle with complex entities and tagging long sequences \citep{dai2018recognizing}.%than very short sequences.
%\citep{dai2018recognizing}.

Although our model results did not surpass the baselines  \citep{bernsohn2024legallens}, further exploration of medium and large variants of GLiNER could be beneficial. Due to limited computational resources, we were unable to include them in our experiments.

%In Table , we see that the bi-encoder model has a better performance over the original GLiNER  and the poly-encoder model. In experiments, we found that shuffling entity order and randomly dropping entities did not affect the metrics. 

%Once we found the best model, we also trained this model on a synthetic dataset that was generated using few-shot learning. Prompts and examples in the Appendix \ref{appendix} However, no improvement was found in the results.\footnote{Synthetic dataset was not validated.}

%Once we found the best model, we applied rule based entity filtering that improved the results on the test dataset by 0.5\% increasing the final F1 score to 60.01\% from 59.53\% the 

%The distribution of number of words by entity type is shown in Figure \ref{fig:number_of_words}. The entity type VIOLAION has an average number of 12 words as compared as to 4 for other three entity types. This increases the complexity of this task. The model performs well on the later set of entities compared to later as shown in Table \ref{tab:NER_by_ents}.
%It has been shown in prior research works that  NER models do not not with complex entities \citep{dai2018recognizing}.

%Although, our model did not exceed the performance of the baseline models in \citep{bernsohn2024legallens}, GLiNER medium and large can further be explored. Due to lack of compute resources, we could not include these models as a part of our experimentation.

\begin{figure}
    \includegraphics[width=\columnwidth]{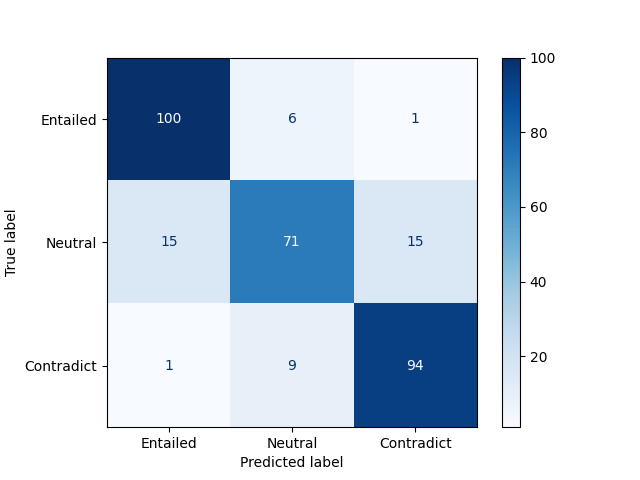}
    \caption{Final Confusion Matrix on the LegalLens NLI the dev dataset }
    \label{fig:cm}

\end{figure}
\paragraph{NLI}
For each legal type category, we employ four distinct models. During the evaluation process on the unlabeled test set, we consider the collective assessment of all four models. The final label for a premise-hypothesis pair is determined by the model exhibiting the highest confidence score among the four.
Our findings indicate that data augmentation proved beneficial, albeit to a certain extent. When we expanded the dataset to triple its original size by incorporating an additional set of paraphrases, we observed that the corresponding increase in F1 scores was not proportional to the increase in data volume. This suggests that there may be diminishing returns in terms of performance improvement beyond a certain threshold of data augmentation.

%In the error analysis of the final model, we do not see the first-class errors---confusions between  \texttt{Contradict} and \texttt{Entailed}---similar to those shown by \citet{bernsohn2024legallens}. We observe a lower number of second-class errors---misclassification of  \texttt{Contradict} or \texttt{Entailed} as \texttt{Neutral}--- than those shown by \citet{bernsohn2024legallens}. However, we observe that a much higher number of \texttt{Neutral} claims being misclassified as \texttt{Contradict} or \texttt{Entailed} as shown in Figure \ref{fig:cm}.

We compare our results with the baseline proposed by \citet{bernsohn2024legallens}. They finetune Falcon 7B \citep{almazrouei2023falcon} and report an F1 score of 81.02\%  compared to 84.73\% for our model. The system ranked fifth on the leaderboad with an F1 score of 65.30\% on the test dataset \citep{hagag2024legallenssharedtask2024}.

In the error analysis of the final model, we see that both the models are capable of handling first class errors---confusions between  \texttt{Contradict} and \texttt{Entailed}. However, our model does better with handling second-class errors---misclassification of  \texttt{Contradict} or \texttt{Entailed} as \texttt{Neutral}; and  Falcon 7B model does better with handling another class of errors---misclassification of   \texttt{Neutral} as \texttt{Contradict} or \texttt{Entailed}. The confusion matrix for our model is shown in Figure \ref{fig:cm}.
%we do not see the first-class errors---confusions between  \texttt{Contradict} and \texttt{Entailed}---similar to those shown by \citet{bernsohn2024legallens}. We observe a lower number of second-class errors---misclassification of  \texttt{Contradict} or \texttt{Entailed} as \texttt{Neutral}--- than those shown by \citet{bernsohn2024legallens}. However, 

It is interesting to note that a multitask DeBERTa based encoder surpassed the performance of a 7B parameter by 3.17\%. Our model is capable of resolving the ambiguities and complexities related to wage norms. Finally, it can be stated that paraphrasing can serve as a data augmentation technique to enhance the natural language understanding capabilities of smaller models.

\section{Conclusion}
\label{conclusion}
%Both NER and NLI models outperform the LLM baselines, as noted by \citet{bernsohn2024legallens}.

In conclusion, we present two systems developed for the LegalLens 2024 shared task, comprising a zero-shot bidirectional DeBERTa encoder with domain-adaptive pretraining for the NER subtask and a multitask DeBERTa encoder enhanced by data augmentation techniques for the NLI subtask. The experiments demonstrate that synthetic data generation can enrich datasets and improve the performance of encoder-based models. However, it is evident that more data does not necessarily translate to better performance. By optimizing on smaller but richer datasets and employing suitable training objectives, smaller models can outperform larger language models.

Further exploration of different augmentation strategies, with a particular focus on generating more contextually diverse synthetic data, employing adversarial data, or leveraging domain-specific paraphrasing techniques, may yield performance improvements for NER tasks. While rule-based filtering improved the F1 score by 0.5\%, the adoption of more sophisticated post-processing strategies, such as probabilistic methods or ensemble techniques, holds the potential to further enhance the results.

Finally, the proposed systems secured the sixth and fifth ranks in the LegalLens NER and LegalLens NLI tasks, respectively, demonstrating their competitiveness in the shared task.%Both NER and NLI models outperfrom the LLM baselines as mentioned by .  Synthetic data generation can enrich data and improve the encoder based lightweight models. However, on the NLI task, we saw that more data doesn't necesarily mean better performance. By optimizing on small but rich dataset and using suitable training objectives, even smaller models can give better performance than LLMs

\section{Acknowledgement}
All experiments were carried out using Kaggle notebooks, which were equipped with a single NVIDIA P100 GPU. We extend our gratitude to Kaggle for their support in providing this computational resource. %The powerful GPU enabled us to efficiently process and analyze the datasets, significantly accelerating our research and allowing for more comprehensive experimentation.

\bibliography{custom}

\appendix

\section{Example Appendix}

\label{appendix}

\begin{figure*}[ht]
    \centering
    \includegraphics[width=0.9\linewidth]{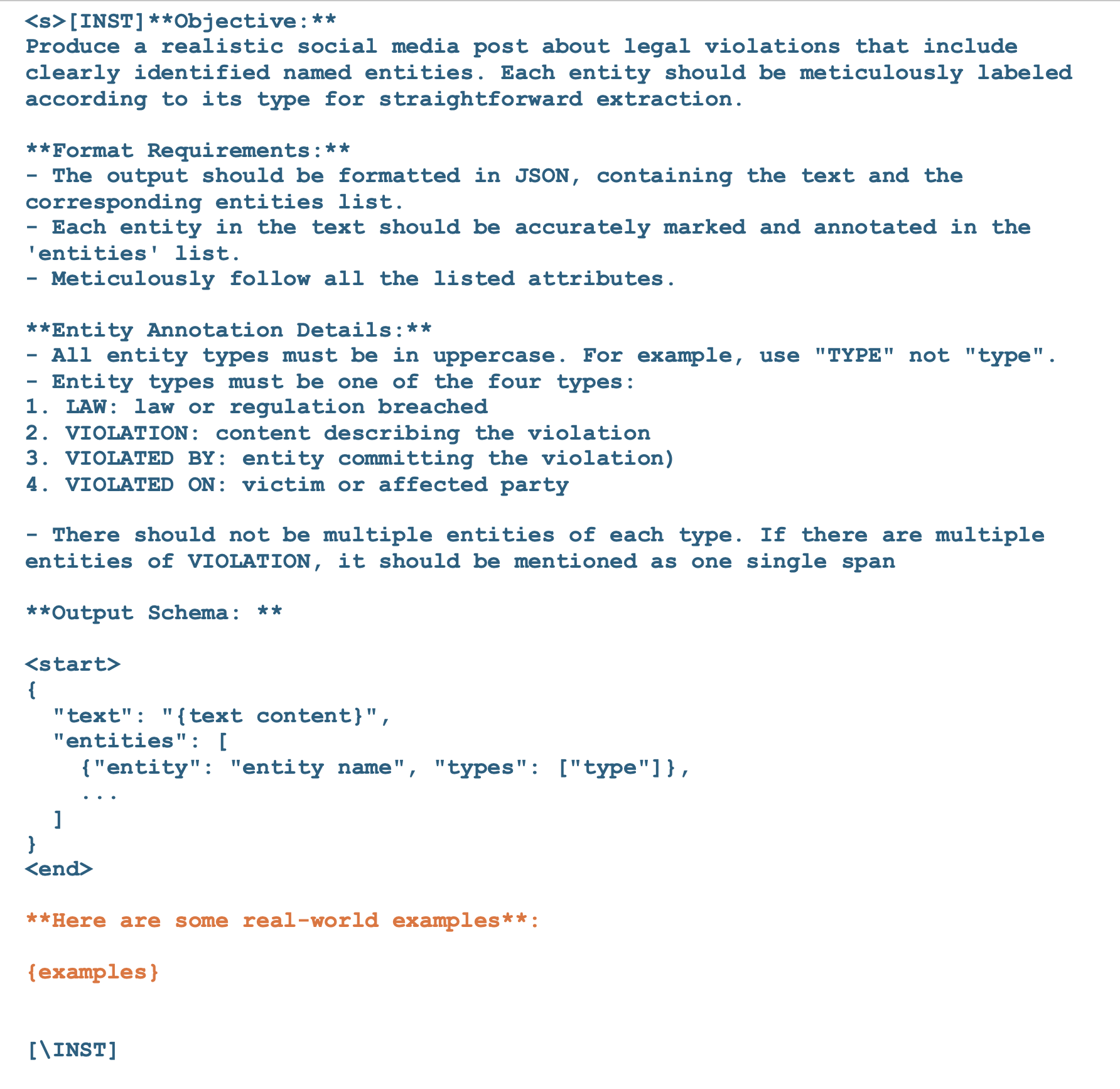}
        \caption{Prompt design for NER dataset with task description and few-shot examples}
\label{nerprompts}
\end{figure*}

\paragraph{Prompts} Figure \ref{nerprompts} showcases few-shot learning approach to generate NER data points using three randomly selected examples from the training dataset.

\begin{figure*}[ht]
    \centering
\includegraphics[width=0.9\linewidth]{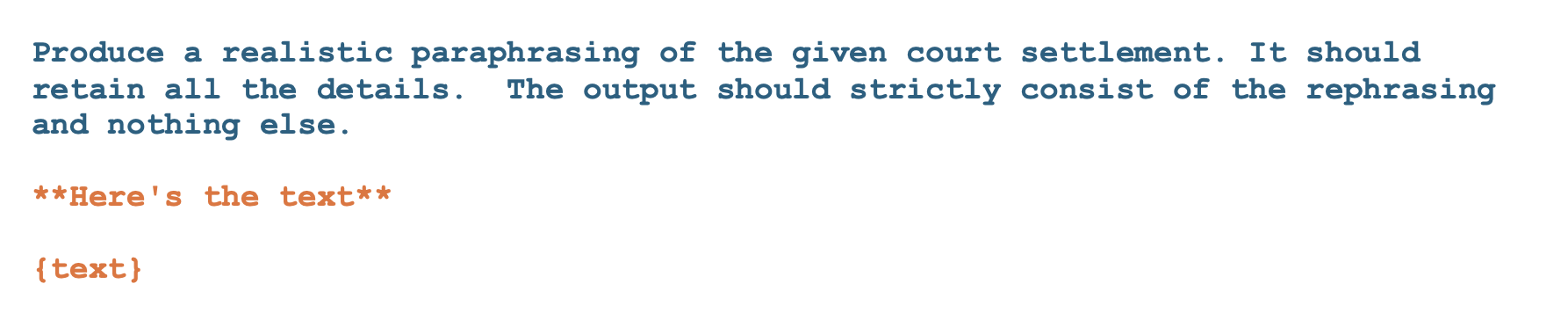} 
 \caption{Prompt design to paraphrase the premise of the NLI training dataset.}

\label{nli_prompts_pre}
\end{figure*}

\begin{figure*}[ht]
    \centering
    \includegraphics[width=0.9\linewidth]{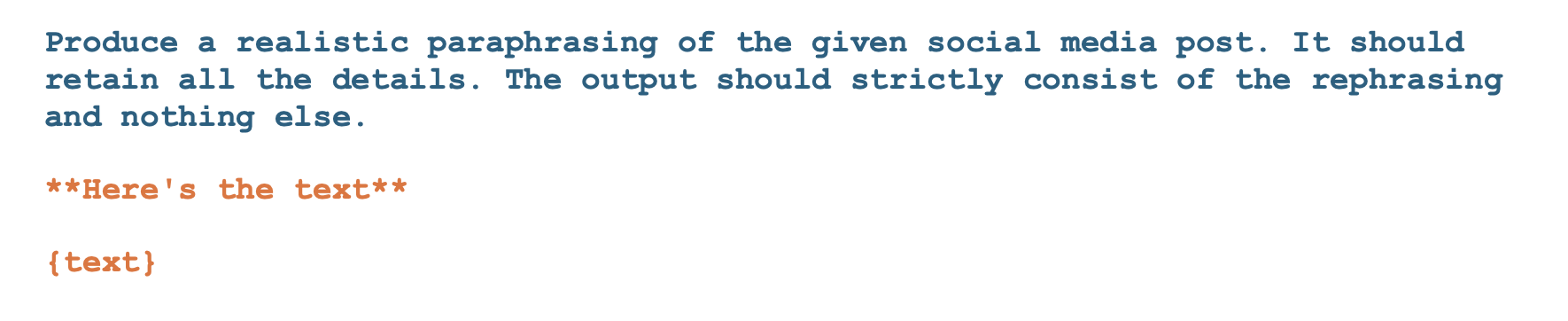}
    \caption{Prompt design to paraphrase the hypothesis of the NLI training dataset.}
\label{nli_prompts_hyp}
\end{figure*}

Figure \ref{nli_prompts_pre} and \ref{nli_prompts_hyp} showcase prompts to generate praphrases of premise and hypothesis of the NLI training dataset.

% \begin{figure*}[ht]
%     \centering
%     \begin{subfigure}{\textwidth}
%         \centering
%         \includegraphics[width=0.9\linewidth]{prompt_nli_hypothesis.png}
%         \caption{Prompt design for Implicit NER data set. Prompt contains the \textcolor{blue}{task description}, \textcolor{orange}{few-shot examples}, and \textcolor{green}{specific instructions}.}
%     \end{subfigure}
%    \begin{subfigure}{\textwidth}
%     \centering
%     \includegraphics[width=0.9\linewidth]{prompt_nli_premise.png}
%     \caption{Prompt design for Explicit NER data set. Prompt contains the \textcolor{blue}{task description}, \textcolor{orange}{few-shot examples}, and \textcolor{green}{specific instructions}.}
%     \end{subfigure}
%     \caption{The prompts used for generating the NER data set.}
% \label{nli_prompts}
% \end{figure*}

\end{document}